\documentclass[10pt,twocolumn,letterpaper]{article}

\usepackage[pagenumbers]{iccv} 

\usepackage[dvipsnames]{xcolor}
\usepackage{amsmath}
\usepackage{amssymb}
\usepackage{subcaption}
\usepackage{xcolor}
\usepackage{resizegather}

\def\realnum{\mathbb{R}}
\def\image{\mathcal{I}}
\def\video{\mathcal{V}}

\def\feature{\mathbf{F}}
\def\triplane{\mathbf{\textit{T}}}
\def\loss{\mathcal{L}}
\def\expectation{\mathbb{E}}

\def\token{\mathit{t}}
\def\augview{\image^\mathcal{A}}
\def\randview{\image^\mathcal{S}}

\definecolor{iccvblue}{rgb}{0.21,0.49,0.74}
\usepackage[pagebackref,breaklinks,colorlinks,allcolors=iccvblue]{hyperref}

\def\paperID{***} 
\def\confName{ICCV}
\def\confYear{2025}

\title{UVRM: A Scalable 3D Reconstruction Model from Unposed Videos}

\makeatletter
\def\@fnsymbol#1{\ensuremath{\ifcase#1\or \dagger\or \ddagger\or
   \mathsection\or \mathparagraph\or \|\or **\or \dagger\dagger
   \or \ddagger\ddagger \else\@ctrerr\fi}}
\makeatother
\author{
Shiu-hong Kao$^{1,2,}$\thanks{Work mainly done during internship at Microsoft Research Asia.} \quad Xiao Li$^{1}$ \quad Jinglu Wang$^{1}$ \quad Yang Li$^{1}$\\ \quad Chi-Keung Tang$^{2}$ \quad Yu-Wing Tai$^3$ \quad Yan Lu$^{1}$ \vspace{0.3em} \\
{\normalsize $^1$Microsoft Research Asia} \quad
{\normalsize $^2$Hong Kong University of Science and Technology} \quad 
{\normalsize $^3$Dartmouth College}\\
{\tt\small \{skao, cktang\}@cse.ust.hk,} \quad {\tt\small yu-wing.tai@dartmouth.edu,} \quad \\
{\tt\small \{li.xiao, jinglu.wang, v-yangli8, yanlu\}@microsoft.com} 
}

\begin{document}
\maketitle
\begin{abstract}
Large Reconstruction Models (LRMs) have recently become a popular method for creating 3D foundational models. Training 3D reconstruction models with 2D visual data traditionally requires prior knowledge of camera poses for the training samples, a process that is both time-consuming and prone to errors.
Consequently, 3D reconstruction training has been confined to either synthetic 3D datasets or small-scale datasets with annotated poses.
In this study, we investigate the feasibility of 3D reconstruction using unposed video data of various objects. 
We introduce \textbf{UVRM}, a novel 3D reconstruction model capable of being trained and evaluated on monocular videos without requiring any information about the pose.
UVRM uses a transformer network to implicitly aggregate video frames into a pose-invariant latent feature space, which is then decoded into a tri-plane 3D representation.
To obviate the need for ground-truth pose annotations during training, UVRM employs a combination of the score distillation sampling (SDS) method and an analysis-by-synthesis approach, progressively synthesizing pseudo novel-views using a pre-trained diffusion model.
We qualitatively and quantitatively evaluate UVRM's performance on the G-Objaverse and CO3D datasets without relying on pose information. Extensive experiments show that UVRM is capable of effectively and efficiently reconstructing a wide range of 3D objects from unposed videos. 
\vspace{-0.1in}
\end{abstract}    
\section{Introduction}\label{sec:intro}
The task of digitally reproducing, modifying, and photo-realistically rendering 3D scenes and objects stands as a core research area in computer vision, with wide-ranging applications. 
As digital landscapes and interactive technologies increasingly pervade sectors such as entertainment, robotics, and design, the demand for scalable and adaptable 3D models is at an all-time high.
Recent breakthroughs in neural 3D representations, initiated by NeRF~\cite{mildenhall2021nerf} and expanded upon by subsequent methods like 3DGS~\cite{kerbl20233d} have achieved unprecedented results, paving the way for the creation of generalized 3D foundation models.
These models, analogous to the progress in natural language processing and text-to-image models in computer vision, will enable more efficient and flexible 3D content generation, manipulation, and interaction at scale. 

\begin{figure}
    \centering
    \includegraphics[width=\linewidth]{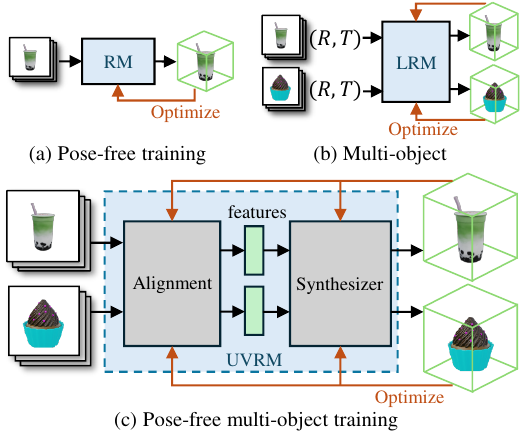}
    \caption{Different from previous methods, which either focus on (a) per-scene pose-free training or (b) 3D reconstruction model trained with known camera poses, we propose UVRM (c) aims for a fully pose-free training of 3D reconstruction model from 2D observations.}
    \label{fig:teaser}
    \vspace{-0.1in}
\end{figure}
Large Reconstruction Models (LRMs) have contributed notable development on 3D foundation models, offering the ability to reconstruct 3D representations from various objects or scenes in a single forward pass. Currently, LRMs are trained on synthetic 3D datasets, like Objaverse~\cite{objaverse} and its derivatives.
Although effective for now, the reliance on synthetic 3D data is increasingly seen as a bottleneck due to the upper-bound quantity and quality of such data, especially in comparison to the abundant data available in image, video, and speech domains.
To overcome this, alternative approaches for 3D reconstruction training use large video and image datasets.
However, these approaches require precise camera poses to align 3D representations with their corresponding 2D observations for training using reconstruction losses such as render loss.
The scarcity of multi-view datasets with accurate pose annotations, combined with the challenges of estimating camera poses from arbitrary images or videos, due to issues like homogeneous regions or view-dependent appearances, poses a significant barrier. Thus, creating a pose-free training and inference framework for 3D reconstruction is crucial for scaling 3D reconstruction efficiently and advancing toward the goal of building generalized 3D foundation models.

In this paper, we take an important step toward harnessing large-scale 2D datasets for 3D reconstruction. Our primary objective is to explore whether it is feasible to train a 3D reconstruction model for various objects, using only monocular videos without any pose annotations.
We tackle this inquiry from two angles:
\begin{itemize}
    \item \textbf{Pose-free alignment} - \textit{Is it possible to achieve 3D reconstruction from an arbitrary number of input views without explicit pose alignment?}
    \item \textbf{Pose-free training} - \textit{Can we train a reconstruction model solely with 2D data, devoid of pose annotations?}
\end{itemize}

We affirmatively answer these questions by introducing UVRM, a \textit{\textbf{R}}econstruction \textit{\textbf{M}}odel for \textit{\textbf{U}}nposed \textit{\textbf{V}}ideos.
To tackle the first question, we propose encoding 3D scenes using viewpoint-invariant features. This differs from traditional approaches, which typically focus on estimating view-specific or pixel-aligned features with explicit pose calibration.
Our method encodes all input views into a unified latent space, utilizing a transformer-based model to implicitly aggregate information from multiple views.
The result is a latent feature with viewpoint-invariant tokens. This strategy allows for the scaling of inputs to dense, pose-free multi-view images without compromising on memory or computational efficiency.
These viewpoint-invariant tokens can then be decoded into a neural 3D representation suitable for rendering.
To address the second question, we integrate the Score Distillation Sampling (SDS) method with an analysis-by-synthesis strategy. This involves incrementally augmenting view-consistent pseudo-views using pre-trained diffusion models throughout the training process. 
Our method circumvents the need to calculate render loss between input and reconstructed views during training, thereby obviating the necessity for ground truth pose annotations for input videos.

We evaluate our model's performance on the synthetic G-Objaverse dataset~\cite{zuo2024sparse3d} without using of pose information, as well as CO3D~\cite{reizenstein21co3d} dataset with real-world videos.
We demonstrate that the proposed UVRM can reconstruct various 3D objects from pose-free monocular videos. 
Notably, UVRM outperforms prior pose-free NeRF methods that rely on per-object optimization, showcasing superior results. Our contributions can be summarized as:
\begin{itemize}
    \item A new research problem of training 3D reconstruction model from 2D datasets without explicit pose calibration.
    \item A new method that takes pose-free monocular videos as input for 3D object reconstruction.
    \item A novel training pipeline that eliminates pose annotations for training 3D reconstruction models.
\end{itemize}
\vspace{-.1in}

\section{Related Work}\label{sec:related}
\noindent{\bf Neural 3D Representations.}
The neural field representation of 3D scenes has attracted significant attention from the literature since the pioneer work of NeRF~\cite{mildenhall2021nerf}.
NeRF has demonstrated its effectiveness on the task of view synthesis from multi-view posed images, leading to a number of follow-up works that extend its capabilities.
Some representative works including NeUS~\cite{wang2021neus}, Tri-plane~\cite{chan2022efficient}, and Gaussian splatting~\cite{kerbl20233d}.
These techniques utilize multi-layer perceptrons to generate implicit or hybrid fields such as sign distance functions or volume radiance fields.
Our method leverage the expressiveness of neural representations, aims to reconstruct 3D objects with minimal requirement of inputs (i.e., pose-free monocular video).

\vspace{0.2em}
\noindent{\bf Multi-view 3D Reconstruction.}
Vanilla multi-view reconstruction method, regardless of its 3D representations, requires accurate camera poses. Camera poses are often obtained from Structure-from-Motion (SfM) methods such as COLMAP~\cite{schonberger2016structure}, which significantly increases the time cost and risk of failure, due to the sensitivity of traditional feature matching strategy. 
Some recent works incorporates this pipeline with neural representations to jointly improve the reconstruction quality and camera estimation robustness~\cite{inerf,wang2021nerf,xia2022sinerf,fu2023colmapfree,fan2024instantsplat}.
Another stream of works rely on additional information such as depth~\cite{bian2023nope,sun2024correspondence} or optical flow~\cite{meuleman2023progressively}.
Our method focus on 3D reconstruction from pose-free videos without explicit pose calibration.

\begin{figure*}
    \centering
    \includegraphics[width=\linewidth]{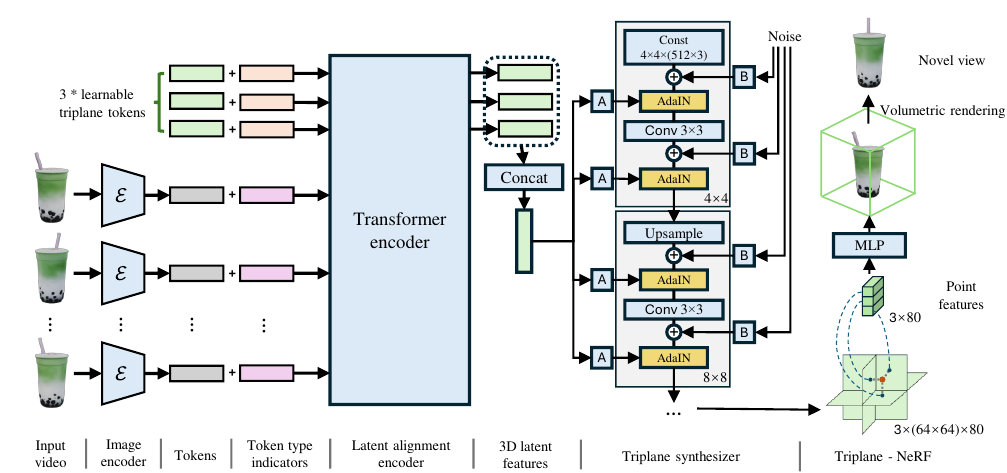}
    \caption{\textbf{UVRM architecture.} We propose UVRM, a transformer-based reconstruction model for pose-free monocular video inputs. It first encodes each input view into latent space with a VAE encoder~\cite{vae}. Next, it adopts a T5-based transformer encoder~\cite{2020t5} to extract a pose-invariant feature by implicitly aligning the image latent sequence. The extracted feature are then used modulate a style-based~\cite{karras2019style} synthesizer to output a tri-plane representation. Here ``A'' implies a learned affine transform, and ``B'' stands for learned per-channel scaling factors to the noise input.}
    \label{fig:overview}
    \vspace{-0.2in}
\end{figure*}

\vspace{0.2em}
\noindent{\bf Large Reconstruction Models.} Considering the expensive time complexity of per-scene optimization, many works have proposed training a large reconstruction model for multiple objects or scenes. These methods directly reconstruct the 3D representation in a feed-forward pass.
Early approaches directly train a CNN-based network to predict neural points~\cite{wiles2020synsin} or multi-plane images (MPIs)~\cite{zhou2018stereo}, synthesizing novel views via rendering methods such as alpha compositing or point splatting. 
PixelNeRF~\cite{yu2021pixelnerf} predicts pixel-aligned features from images for conditional radiance field. Following works improve the performance of multi-scene model using feature matching~\cite{chen2023explicit,charatan23pixelsplat,wewer24latentsplat,chen2024mvsplat}, geometry-aware attention~\cite{miyato2023gta,li2024mlrm}, large transformer backbone~\cite{hong2023lrm,zhang2024gslrm,pflrm}, or 3D volume representations~\cite{xu2024murf}. 
We do not directly train a LRM in this paper; instead, we focus on developing a new technology for reconstructing 3D objects from pose-free videos, offering novel insights for future expansion of LRM training to large-scale visual datasets.
\section{Method}\label{sec:method}
We aim to train a 3D reconstruction model (i.e., the UVRM) from collections of unposed video sequences, each representing one object from different views. 
Our solution (\cref{fig:overview}) consists of two key components: a pose-free, multi-view transformer that implicitly aligns an arbitrary number of RGB frame sequence into a pose-invariant latent feature (Subsection~\ref{uvrm}), and a novel training framework that eliminates the usage of ground-truth poses to compute the render loss for reconstruction (Subsection~\ref{training_method}). 
Our framework is build upon the recent advance of neural 3D representations and a diffusion prior with score distillation sampling. We will therefore first introduce some preliminaries in Subsection~\ref{preliminary}. Then, we introduce the pose-free alignment and the pose-free training in following subsections.


\subsection{Preliminaries}\label{preliminary} 

\noindent{\bf Triplane NeRF.} 
A Neural Radiance Field (NeRF) is a 5D function that represents the volumetric radiance of any 3D objects~\cite{mildenhall2021nerf}.
The vanilla NeRF adopts a fully implicit approach, querying the radiance at each position with a MLP network.
A triplane is an hybrid NeRF representation for 3D objects~\cite{chan2022efficient,gao2022get3d}, which is composed of three axis-aligned feature planes $\triplane=(\triplane_{XY}, \triplane_{XZ}, \triplane_{YZ})$, each with the dimension of $H\times W \times d_\triplane$, where $H\times W$ is the spatial resolution and $d_\triplane$ is the number of feature channels.
Triplane queries the radiance value by first projected 3D positions onto each of the axis-aligned plane and query the corresponding point features $\hat{\triplane}\in\realnum^{3\times d_\triplane}$. These features are then decoded into color and density via a smaller MLP. We adopt the tri-plane NeRF as our neural representation for 3D objects.


\vspace{0.5em}
\noindent{\bf Score Distillation Sampling (SDS).} 
SDS is 
a powerful loss fuction for training 3D generative models from text~\cite{poole2022dreamfusion,tang2023dreamgaussian}, which
can be regarded as a prior that maximizes the agreement of the rendered image from some 3D representations with given text condition. It works by providing gradients towards image formed through the conditional denoising process of a pre-trained diffusion model applied on the rendered image~\cite{alldieck2024score}.
Formally, given a parametric 3D representation $g_\Theta$ and a random camera pose $p_r$, the SDS loss can be written as:
\begin{gather}
    \image=g_\Theta(p_r) \\
    \image_t=\sqrt{\alpha_t}\image + \sqrt{1-\alpha_t}\epsilon \\
    \nabla_\Theta\loss_{SDS}=\expectation_{t,p_r,\epsilon}\left[w\left(t\right)\left(\epsilon_\phi\left(\image_t;t,e\right)-\epsilon\right)\frac{\partial\image}{\partial\Theta}\right] \label{eq:sds}   
\end{gather}
where $w(\cdot)$ is a weighting function for denoising timestep $t$, $\epsilon\sim\mathcal{N}(0,1)$ is a random Gaussian noise, $\epsilon_\phi(\cdot)$ is the noise predicting function, i.e., the pre-trained diffusion network with parameters $\phi$, and $e$ is the given text embedding.
We show (in later section) that the SDS loss can be adopted as a weak-supervised loss for multi-view reconstruction from pose-free video frames.



\subsection{UVRM Architecture}
\label{uvrm}
Our UVRM model takes an arbitrary numbers of video frames as input and produces a tri-plane that represents the object in the video.
It consists of three components: an image encoder, a latent alignment encoder, and a triplane synthesizer. 

\vspace{0.5em}
\noindent{\bf Image encoder.} We utilize the encoder of a pretrained VAE from stable diffusion~\cite{rombach2021highresolution}, which projects a given video $\video\in\realnum^{H\times W \times 3}$ into a latent $\hat{\ell}\in\realnum^{h\times w\times d}$ with smaller resolution.
Each frame in the video is encoded and flattened, forming a token sequence $\ell\in\realnum^{N\times d'}$ where $N$ is the number of frames in the input video and $d'=h\times w\times d$.

\vspace{0.5em}
\noindent{\bf Latent alignment encoder.} The pre-processed token sequence $\ell$ can be of arbitrary length, with each token representing the object in arbitrary and unknown poses. Before reconstructing the 3D representation that is applicable for rendering, we first leverage a transformer model to compress the token sequence into a fixed number of tokens.
During the compression training, the transformer model eventually learns to ignore obstructions and focused on implicitly aligns different input tokens of pose-free view observations into these unified, fixed number of tokens, forming a complete latent representation of the 3D object in the video. 
We implement this process by using three learnable tokens $\token\in\realnum^{3\times d'}$ representing the 3D latent feature. The information of input tokens are compressed into $\token$ using a transformer encoder $\mathcal{T}$~\cite{vaswani2017attention} , by prefixing the 3D latent $\token$ to the video token sequence $\ell$ as prompts:
\begin{equation}
    \hat{\token}, ...=\mathcal{T}(\token+\token^\triplane, \ell+\token^V),
\end{equation}
where $\token^V\in\realnum^{d'}$ and $\token^\triplane\in\realnum^{d'}$ are token type indicators.
We only extracts the first three tokens in the output sequences corresponding to compressed 3D latent and concatenate them as the 3D latent features $\feature\in\realnum^{(3\times d')}$.

\vspace{0.5em}
\noindent{\bf Triplane synthesizer.}
We use a style-based triplane synthesizer~\cite{karras2019style} (\cref{fig:overview}) to decode the 3D latent features into a triplane. Similar to~\cite{karras2019style}, the triplane synthesizer progressively decodes 
a set of learnable feature maps into a triplane feature $\triplane \in \realnum^{3\times H_\triplane \times W_\triplane \times d_\triplane}$ with stacked convolution layers. The encoded 3d latent token $\feature$ modulates each convolution layer output $\mathbf{x}_i$ with an adaptive instance normalization (AdaIN) layer:
\begin{gather}
    (\mathbf{y}_{s, i},\mathbf{y}_{b, i}) = \mathbf{A}_i(\feature) \\
    \text{AdaIN}
    (\mathbf{x}_i,\mathbf{y})=\mathbf{y}_{s,i} \frac{\mathbf{x}_i-\mu(\mathbf{x}_i)}{\sigma(\mathbf{x}_i)}+\mathbf{y}_{b,i},
\end{gather}
where each $\mathbf{A}_i$ is a learnable affine transformation layer.
We then following the volumetric rendering method in~\cite{chan2022efficient} to query the radiance field for rendering images.

\subsection{Pose-free Training}
\label{training_method}

The UVRM we introduced in \cref{uvrm} has eliminated the requirement of explicit camera calibration for video inputs. However, existing pipelines for 3D reconstruction, still require pose annotations during training. 
Pose annotations are used to render the 3D representation into reference views w.r.t. the input video for computing reconstruction loss.
In contrast, we propose a pose-free training pipeline which combines both \textbf{weak supervision} and \textbf{self-supervision}. Specifically, we \textbf{weakly supervise} the training with the SDS loss, and \textbf{self-supervise} the training using pseudo novel views that are generated from the model itself.

\paragraph{Weak-supervision with SDS loss.}
For multi-view reconstruction, we replacing the condition $e$ of the SDS loss with input video $\video_{gt} = \{\image_{gt}\}$:
\begin{equation}\label{eq:sds_v} 
\nabla_\Theta\loss_{SDS}=\expectation_{t,p_r,\epsilon}\left[w\left(t\right)\left(\epsilon_\phi\left(\image_t(p_r, \Theta);t,\image(p_{gt})\right)-\epsilon\right)\frac{\partial\image}{\partial\Theta}\right]
\end{equation}
where $\epsilon_\phi$ is the noise prediction network $p_r$ is random sample camera pose for rendering, and $p_{gt}$ is the ground truth pose (unknown for us) of input image $\image_{gt}$.

A good property of \cref{eq:sds_v} is that it
attempts to match the distribution of images generated from 3D representations $P(\image|\Theta, p_r)$ to the distribution of ground truth images $P(\image|p_{gt})$, up to an global affine transformation of the camera system (i.e., the matched distribution preserves the relative camera pose between different views).
Consequentially, we do not need to access the ground truth pose $p_{gt}$ for computing the loss function.

In practice, we compute \cref{eq:sds_v} stochastically with a pre-trained image-to-3d diffusion model~\cite{liu2023zero1to3}.
In each training step, we randomly render a small number of $k$ views $\{\randview_i\}$ to estimate the gradient of SDS loss w.r.t a reference image $\image^R$ from the video.
The random pose ${p}$ for $\{\randview_i\}$ for rendering is sampled as follows:
\begin{equation}
    \left\{p_i=\left(r,\frac{2\pi}{i},\delta\right)\right\}_{i=1}^{k}\gets P(k;r,\theta),
    \label{eq:pose_sample}
\end{equation}
where $\delta\sim U(-\theta,\theta)$, and $k,r,\theta$ are all hyper-parameters with $k\in\mathbb{N}$, $r>0$, $\theta\in[0,\pi/2]$. This generator samples $s$ camera poses orbiting the center of coordinate system with radius $r$, fixed azimuth angles, and random polar angles.

One issue of the SDS loss (\cref{eq:sds_v}) is that a perfect match is only theoretically guaranteed when the randomly sampled pose distribution $p_r$ for rendering the 3D representation matches the unknown, ground truth camera distribution $p_{gt}$ in the video, which is hardly the case when handling real-world videos.
The stochastic approximation of the SDS loss also introduces additional uncertainty for the optimization process.
Hence, we can only regard the SDS loss as a weak supervision for reconstructing a rough 3D representation.
In the next section we exploit to further improve the training process by combining additional pixel-wise loss with self-supervision.

\vspace{0.5em}
\noindent{\bf Self-supervision by pseudo-view augmentation.}
A naive approach for self-training is to render additional novel views from the UVRM itself.
However, this approach has very limited benefit, as it operates on the  UVRM network trained with SDS loss which only forms an approximation of the desired oracle. 
Our key observation is that re-rendered images of training videos from UVRM (trained with SDS loss) is being optimized towards a specific trajectory, such that each gradient descent step matches a single step of the reversed (i.e., a denoising step) diffusion process.
In simple terms, rendered images from UVRM during the intermediate training stage, can be regarded as a set of partially generated images from the denoising process of the pre-trained diffusion model. 
Hence, we conduct an ``analysis-by-synthesis" approach, by simply reusing the same diffusion model to "take over" rendered image from the partially converged UVRM and perform additional denoising steps to augment these images as new pseudo view for further training.

\begin{figure}
    \centering
    \includegraphics[width=\linewidth]{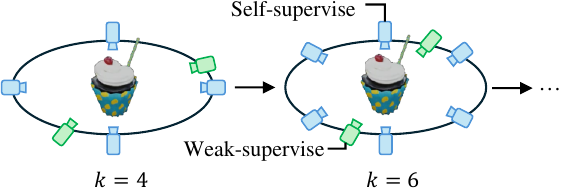}
    \vspace{-0.1in}
    \caption{Illustration of our pose-free training framework, where $k$ pseudo-views are randomly synthesized at scattered poses along a given trajectory to achieve self-supervision and SDS regularization at random poses for weak supervision. We iteratively augment more pseudo-views throughout the training process.}
    \vspace{-0.2in}
    \label{fig:camera}
\end{figure}
Formally, given the partially converged UVRM model $g(\Theta)$, we first render $k$ images from random views sampled from~\cref{eq:pose_sample}:
\begin{equation}
    \{\augview\} = \left\{\left( g_\Theta\left(p_i';V\right), p_i'\right)\right\}_{i=1}^{k'}
\end{equation}
We then add noise perturbation to $\{\augview\}$ following the forward diffusion process, and using the same diffusion model $D$ in \cref{eq:sds} to synthesis augmented images $\{\augview\}'$:
\begin{gather}
\augview_{i,t}=\sqrt{\alpha_{i,t}}\augview_{i,t} + \sqrt{1-\alpha_{i,t}}\epsilon \\
(\augview_i)'=D_\phi\left(\augview_{i,t};t,\image^R\right).
\end{gather}
Finally, we use a combination of the mean square error (MSE) loss and the perceptual loss (LPIPS~\cite{zhang2018unreasonable}) in addition to the SDS loss ($\lambda$ and $\beta$ are loss weights):

\begin{equation}
    \begin{aligned}
        \loss_{\text{recon}}&=\frac{1}{k}\sum_{i=1}^k \text{MSE}\left(g_\Theta\left(p_i';V\right),\augview_i\right) \\
        &+\lambda \cdot\frac{1}{k}\sum_{i=1}^k\text{LPIPS}\left(g_\Theta\left(p_i';V\right),\augview_i\right) \\
        &+\frac{1}{\beta}\loss_{\text{SDS}}.
    \end{aligned} 
\end{equation}

\noindent{\bf Iterative synthesis.}
The rendered image from UVRM in the early stage is optimized with less steps, referring to early steps in the reversed diffusion process.
As the optimization process progresses, the rendered images increasingly resemble the latter stages of the diffusion model's denoising process, requiring fewer additional denoising steps and less noise intensity.
To align with this training process, we introduce an iterative, progressive enhancement strategy that starts with stronger noise $\alpha_{i,t}$, more denoising steps, fewer number of views $k$, and larger weight (lower $\beta$) for the SDS loss in the early stages of training.
During the training stage, we iteratively synthesize new pseudo-views with a fixed interval of training steps, gradually reducing the noise intensity, the number of denoising steps, while decreasing the weight for the SDS loss and augmenting more pseudo views.
Note that our synthesis at the very beginning of the training stage degrades into purely conditional image generation, which corresponds to the reversed denoising process when $t\to+\infty$.

\vspace{0.5em}
\noindent{\bf Discussion.} A straightforward idea for pose-free training is to estimate camera poses for all subjects. We show in later section that indeed UVRM trained with camera poses converges faster and achieves higher quality. Yet this paper aims to propose an orthogonal, new training method different from SfM-based approaches~\cite{schoenberger2016sfm}, by learning from the camera trajectory distribution to produce 3D pose-invariant latent features, akin to and inspired by human's mental 3D reconstruction ability without explicit parameter estimation. 
This idea supports scaling up to larger real-world datasets and bypasses their recurrent problems (e.g., per-scene pose-free training, need dense views, front view only, close poses) in explicit camera estimation. 
To this end, the weak-supervised SDS loss and the self-supervised augmentation, which are complementary to each other, work best when employed together in our pose-free training pipeline (shown in ablation study). 
The diffusion-based augmentation cannot synthesize consistent pseudo-views for training without the SDS loss to drive the partially rendered sample towards the denoising trajectory, while 
the SDS loss itself cannot fully guide the training converged to a high-fidelity solution with sufficient details. Our method works by adapting both supervisions 
such that their complementary effect is maximized during the whole training stage.

\subsection{Model Training}
\label{detals}

\noindent{\bf Model architecture.}
We use the pretrained VAE from Stable Diffusion 2.1~\cite{stablediffusion} as our image encoder.
The latent alignment encoder is a T5~\cite{raffel2020exploring} model with $16$ layers with 8 heads and feedforward dimension of 2048.
 The triplane synthesizer is a StyleGan~\cite{karras2019style} architecture staring from $4\times 4\times 1536$ resolution to $64\times 64 \times 80$ for each triplane.
we use a 4-layer MLP to predict the color and density from triplane features.

\vspace{0.3em}
\noindent{\bf Hyper-parameters.} 
We use the Adam~\cite{kingma2014adam} optimizer, a learning rate of $1e^{-4}$ with 10k warm-up steps and cosine annealing strategy, and a batch size of 4.
$\lambda$ is set to 1.
We linearly increase $\beta$ from $1$ to $25000$. 
We sample 4 random views with $\theta=\frac{\pi}{18}$ to compute SDS in each training step.
We perform iterative augmentation every $6000$ steps with $\theta$ set to 0.
$k$ is set 6 at the beginning and increases by 5 for each augmentation iteration.
\vspace{-0.1em}

\begin{figure}
    \centering

    \includegraphics[width=\linewidth]{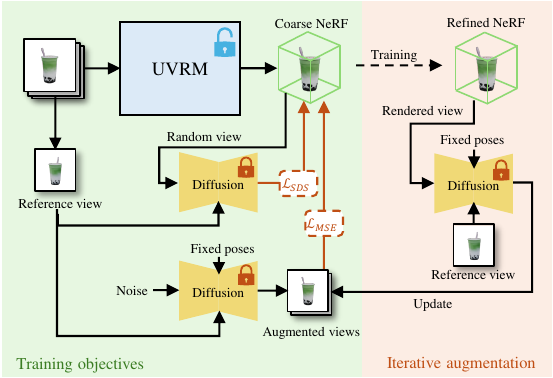}
\vspace{-0.1in}    \caption{\textbf{Iterative augmentation pipeline.} 
We iteratively alternate between (left, green)
weakly supervise training of the UVRM model with score distillation sampling (SDS) on the current set of reference view frames, and (right, orange) generating new set of novel pseudo-views for self-supervised training using the current UVRM and the pre-trained diffusion model. The generated pseudo-views can be trained with pixel-wise render loss.}
    \label{fig:training}
    \vspace{-0.15in}
\end{figure}

\section{Experiments}
To validate our proposed UVRM, we perform an ablation study to demonstrate the ability of our pose-free alignment, the impact of the weak-supervised SDS loss and the self-supervised augmentation strategy. In addition, we perform comparison against two type of existing pose-free methods: an optimization based method for single object and a single image to 3D method.
Finally, we show that our solution works well on real-world video sequences.
\begin{figure}
    \centering
    \includegraphics[width=\linewidth]{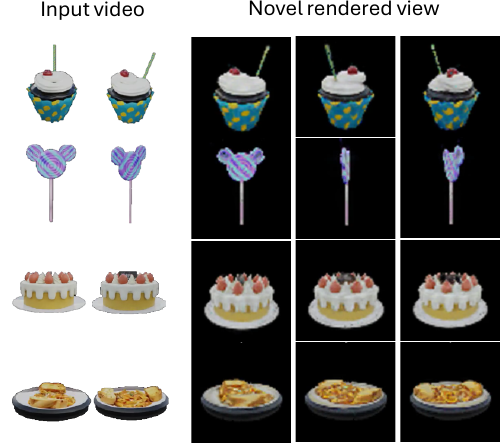}
    \vspace{-0.1in}
    \caption{\textbf{Capability of pose-free alignment.} UVRM is able to conduct pose-free alignment and 3D reconstruction from a set of monocular videos, which shows great potentials in scalability.}
    \label{fig:render_loss}
\end{figure}
\begin{figure*}
    \centering
    \includegraphics[width=.9\linewidth]{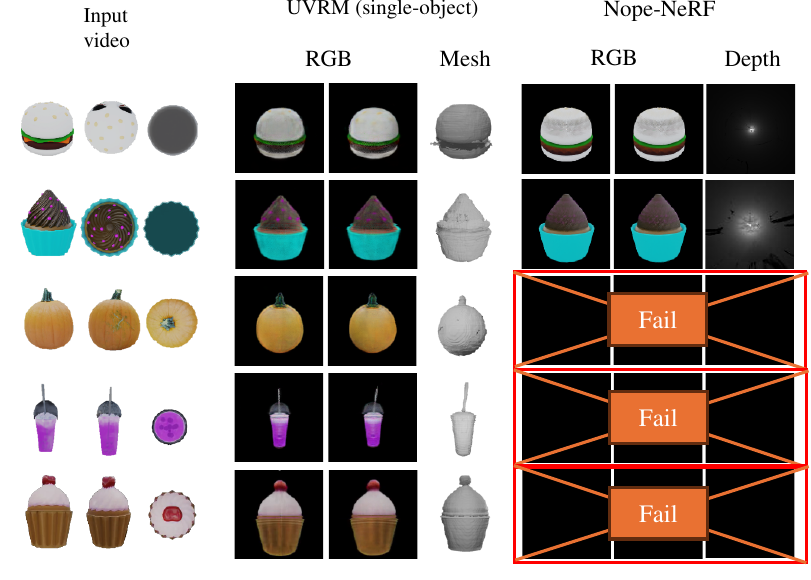}
    \vspace{-0.15in}
    \caption{\textbf{Comparison of pose-free training on single object.} We compare UVRM with Nope-NeRF~\cite{bian2023nope}, the state of the art for NeRF reconstruction with unknown camera poses. In general, UVRM supports more stable reconstruction for 360-degree and sparse views, while Nope-NeRF is shown prone to failure due to the limitation of front view inputs. Depth maps reconstructed from Nope-NeRF also demonstrates a poor reconstructed geometry than ours, while we can directly extract mesh from our UVRM. }
    \vspace{-0.15in}
    \label{fig:single}
\end{figure*}

\begin{figure*}
    \centering
    \includegraphics[width=\linewidth]{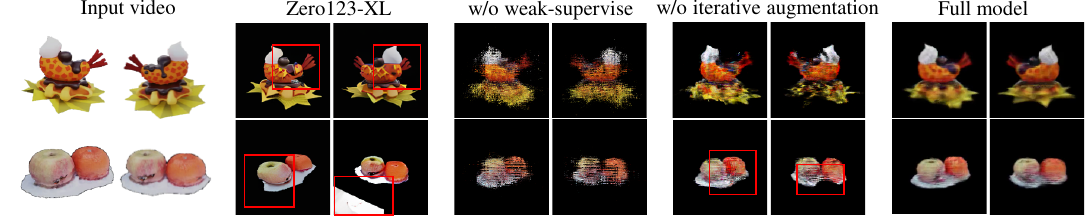}
    \vspace{-0.15in}
    \caption{\textbf{Ablation study.}
    Comparison between the full UVRM, the UVRM without certain components, and Zero123-XL~\cite{liu2023zero1to3}. Specifically, Zero123-XL synthesizes high-frequency image but lacks view consistency.}
    \label{fig:ablation}
    \vspace{-0.15in}
\end{figure*}

\begin{figure}
    \centering
    \includegraphics[width=\linewidth]{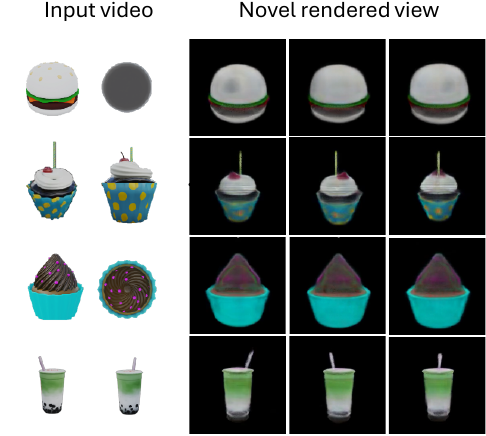}
    \vspace{-0.1in}
    \caption{\textbf{Results of UVRM with pose-free training on video collections from the G-Objeverse dataset.} The first three rows contains subjects that are also in previous experiments (Figure~\ref{fig:render_loss} and~\ref{fig:single}). Overall, UVRM predicts accurate geometry and texture.}
    \label{fig:multiple}
    \vspace{-0.1in}
\end{figure}

\subsection{Experiment Setup}\label{expr_setup}

\noindent{\bf Dataset.}
We use the G-Objaverse Food dataset~\cite{zuo2024sparse3d} in our ablation study and comparisons.
G-Objaverse is a manually annotated subset of the synthetic Objaverse dataset~\cite{objaverse} of ten categories, in which the food category consists of 5314 objects.
Each sample has a video sequence of 40 views with their associated ground-truth pose; We only use the poses for evaluation and discard them during training. 
Our real data experiments are conducted on the CO3D Hydrant dataset~\cite{reizenstein21co3d}.

\vspace{0.3em}
\noindent{\bf Baselines.}
We compare with two pose-free 3D reconstruction methods that most related to ours: a state-of-the-art, optimization based pose-free NeRF method for single object, Nope-NeRF~\cite{bian2023nope}, and a generative image to 3D method (which also serves as our diffusion model used for the SDS loss and view augmentation), Zero123\cite{liu2023zero1to3}.
For baseline methods, we follow their default hyper-parameters and training settings from their official implementation. 
\begin{table}
\centering
\caption{\textbf{Quantitative comparison.} 
We compare the full UVRM method to Zero123-XL and variations of UVRM without our designed components.
}
\vspace{-0.1in}
\resizebox{\linewidth}{!}{%
\begin{tabular}{cccc}
\toprule
    Model & PSNR($\uparrow$)&SSIM($\uparrow$)&LPIPS($\downarrow$)\\\midrule
    Zero123-XL~\cite{liu2023zero1to3} &14.49&0.53&0.23\\
    UVRM (w/o weak-supervise)&15.75&0.69&0.32\\
    UVRM (w/o iterative augmentation)&16.25&0.75&0.24\\
    UVRM (full model)&\colorbox{Salmon!30}{16.54}&\colorbox{Salmon!30}{0.78}&\colorbox{Salmon!30}{0.22}\\\bottomrule
\end{tabular}
}
\vspace{-0.15in}

    \label{tab:recon}
\end{table}

\subsection{Ablation Study}\label{ablation}

\noindent{\bf Pose-free Alignment.} We first validate the capability of the pose-free input alignment of UVRM by training the model on 20 objects randomly sampled from the G-Objaverse Food dataset with known camera poses. To focused on validate the pose-free alignment part without interference, we discard the SDS loss and iterative augmentation and directly use the render MSE loss and LPIPS loss for training.
The reconstructed results are demonstrated in \cref{fig:render_loss}.
Overall, the UVRM is able to reconstruct various detailed 3D objects without pose alignment of input views.

\begin{table}
\centering
\caption{\textbf{Runtime Comparison.} We evaluate the reconstruction time cost with 40 input views on A100 GPU. UVRM significantly speed up the per-object reconstruction time, compared to state-of-the-art pose-free training method.
}\vspace{-0.1in}
\resizebox{\linewidth}{!}{%
\begin{tabular}{ccccc}
\toprule
    Model & \# objects&\# GPUs&Total time (hr.) & Avg. (min/obj.) \\\midrule
    Nope-NeRF~\cite{bian2023nope} &1&1&6.33&380\\
    UVRM (ours)&1&1&16.5&990\\
    UVRM (ours)&20&4&20.65& \colorbox{Salmon!30}{62}\\\bottomrule
\end{tabular}
}
\vspace{-0.15in}

    \label{tab:runtime}
\end{table}

\vspace{0.3em}
\noindent{\bf Pose-free Training.} We train on the same set of 20 objects but this time using our proposed pose-free training pipeline in~\cref{sec:method}.
As shown in Figure~\ref{fig:multiple}, UVRM produces promising results with affordable training complexity (\cref{tab:runtime}). The first three rows provides comparison with those in Fig.~\ref{fig:render_loss} and \ref{fig:single}. The last row showcases a more complicated example with highly non-symmetric geometry, where UVRM also produces reasonable results.

Finally, we show that the weak supervision and  self-supervision used in our pose-free training pipeline is complementary with equally importance.
In \cref{fig:ablation} and in \cref{tab:recon}, results without the self-supervised augmentation maintains the overall 3D structure but struggling to reconstruct further details. Results without the SDS loss, on the other hand, lose the 3D consistency between different views.
\begin{figure}
    \centering
    \includegraphics[width=\linewidth]{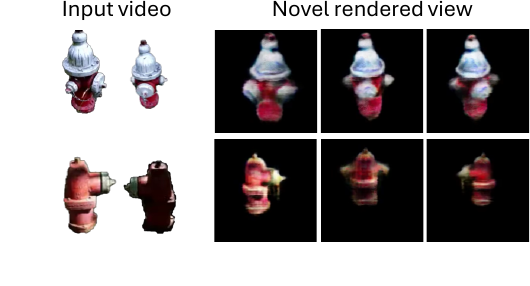}
    \vspace{-0.45in}
    \caption{\textbf{Multi-object results on real-world data}, i.e. CO3D Hydrant~\cite{reizenstein21co3d}. We show that UVRM can not only be trained on multiple objects, but it can also tackle complex objects with non-symmetry and varying shadow.}
    \label{fig:real}
    \vspace{-0.1in}
\end{figure}
\subsection{Comparison}
\label{results}

\noindent{\bf Comparison with pose-free NeRF method.} We compare our proposed method with Nope-NeRF~\cite{bian2023nope}, the state of the art for pose-free NeRF on single object. For fair comparison, we also train UVRM on single object for 50k steps with the iterative training method.
We observe that Nope-NeRF is fragile and fails in many cases (3rd. row to 5th. row in \cref{fig:single}), possibly due to its requirement for dense front views.
Our method is robust under all test cases.
For objects that both UVRM and Nope-NeRF succeeded (1st and 2nd row in \cref{fig:single}), UVRM reconstructs more accurate 3D geometry and textures than Nope-NeRF.
We also compare reconstruction times in \cref{tab:runtime}. Although our method takes longer to fit a single object compared to Nope-NeRF, its robustness and scalability enable the simultaneous reconstruction of multiple objects, significantly reducing the average time required.

\vspace{0.4em}
\noindent{\bf Comparison with single image-to-3D methods.} We also compare our method to the single image-to-3D methods Zero123~\cite{liu2023zero1to3} in \cref{fig:ablation} and \cref{tab:recon}.
While Zero123-XL synthesizes images with high frequency details, it suffers from serious view inconsistency between different views. On the other hand, our UVRMs with weak-supervise training produce more consistent 3D results.


\subsection{Results on Real Videos}
We demonstrate our method's ability to perform 3D reconstruction from collections of real-world sequences without camera pose available, on the CO3D Hydrant dataset.
Real-world videos in the CO3D dataset exhibits larger pose, shape and appearance variations than synthetic data; nevertheless, UVRM reconstructs reasonable results as shown in~\cref{fig:real}. See more results in our supplemental material.

\section{Conclusion}\label{conclusion}
We have proposed a new method, UVRM, for 3D object reconstruction from monocular video collections.
UVRM is a reconstruction pipeline that is fully pose-free: it utilizes a transformer based structure to implicitly align input video frames with arbitrary camera pose, and a novel training method that simultaneously adopts the score distillation sampling method as a weak supervision and an analysis-by-synthesis approach to iteratively augment pseudo-views as self supervision.
We validate our method on both synthetic and real-world datasets without pose information.
Our method takes an important step toward using large-scale 2D datasets for 3D reconstruction.



\vspace{0.3em}
\noindent{\bf Limitations.} 
While our proof-of-concept experiments have demonstrated the possibility of training 3D reconstruction with pose-free 2D data, we have not generalize our method to a large reconstruction model yet, due to limited time budgets and computational resources. 

{
    \small
    \bibliographystyle{ieeenat_fullname}
    \bibliography{main}
}
\clearpage
\appendix
\renewcommand{\thefigure}{A\arabic{figure}}
\renewcommand{\thetable}{A\arabic{table}}
\renewcommand{\theequation}{A\arabic{equation}}
\setcounter{figure}{0}
\setcounter{table}{0}
\setcounter{equation}{0}

\vspace{0.5em}
\begin{center}
    \bf \Large Appendix
\end{center}
\vspace{0.5em}

This document contains additional implementation details for our training and evaluation as well as more results and ablations.
We also provide a video demo for our results in a separate MPEG-4 file along with this document and we encourage readers to watch it.

\section{Implementation Details.}
\noindent{\bf Frame Resolutions.}
In our experiments, all input images are resized to a resolution of $256\times256$. The VAE encoder within the UVRM framework encodes each input frame into a feature of dimension $32\times 32\times 4$. Additionally, the rendering of the tri-plane and the corresponding back-propagation process are performed at a reduced resolution of $64 \times 64$.

\vspace{0.3em}
\noindent{\bf Denoising Timestep Scheduler.}
In the iterative augmentation process, the denoising timestep scheduler is responsible for determining the amount of noise to be added to the rendered image and the number of denoising steps to be performed. As rendered images are progressively optimized through the denoising process, the desired denoising timestep $t$, is dependent on the number of training steps $s$.
The scheduler for the denoising timestep during the iterative augmentation process is defined as follows:
\begin{equation}
t \gets \max(1-\frac{s_{curr}}{s_{total}},0.2) \cdot t_{max}.
\end{equation}
Here, $t_{max}$ represents the maximum number of denoising steps, while $s_{curr}$ and $s_{total}$ denote the current training step and the maximum training step, respectively.
This approach allows for a gradual reduction in the number of denoising timesteps, with a pre-defined minimal threshold.

\vspace{0.3em}
\noindent{\bf Alignment of the Coordinate Systems.}
To evaluate the quantitative performance of UVRM relative to other methods, we align the reference image, $\image^R$, with a predefined camera pose, $p_0$, and utilize the relative pose between the target and reference views for computation.



\vspace{0.4em}

\begin{figure}
    \centering
    \includegraphics[width=\linewidth]{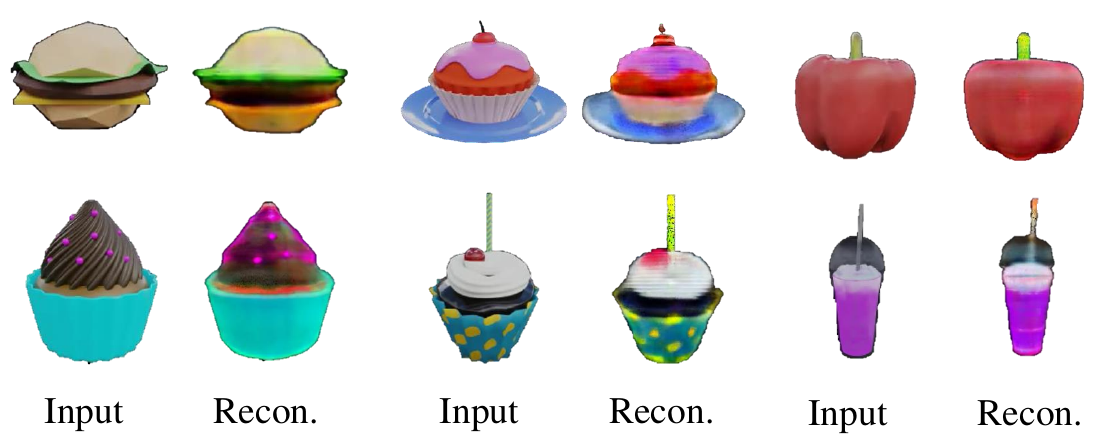}
    \caption{UVRM results on 128 object collections.}
    \label{fig:uvrm_128}
\end{figure}

\vspace{0.4em}
\noindent{\bf Dataset scale.}
\Cref{fig:uvrm_128} demonstrates an additional experiment for our UVRM on 128 object collections. The training process takes around 3 days with 4 A100 GPUs using the same hyperparameters as in the main paper. As shown, UVRM, accompany with the proposed training method, is capable of reconstructing diverse objects concurrently. 

\begin{figure}
    \centering
    \includegraphics[width=0.75\linewidth]{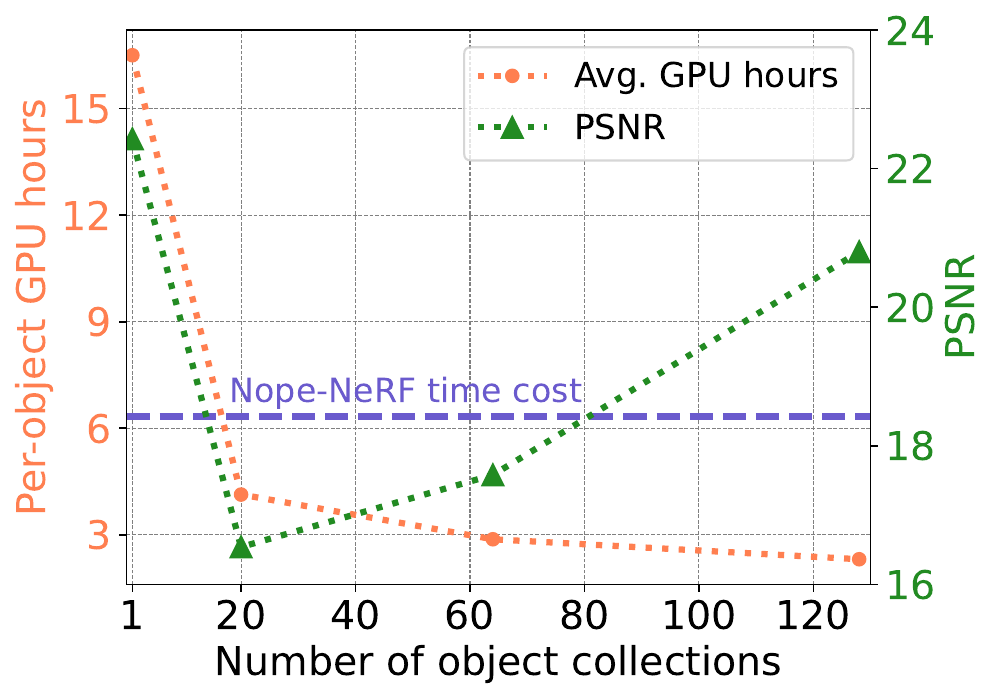}
    \caption{Time cost and quantitative evaluation.}
    \label{fig:time_cost}
\end{figure}
\begin{figure}
    \centering
    \includegraphics[width=.85\linewidth]{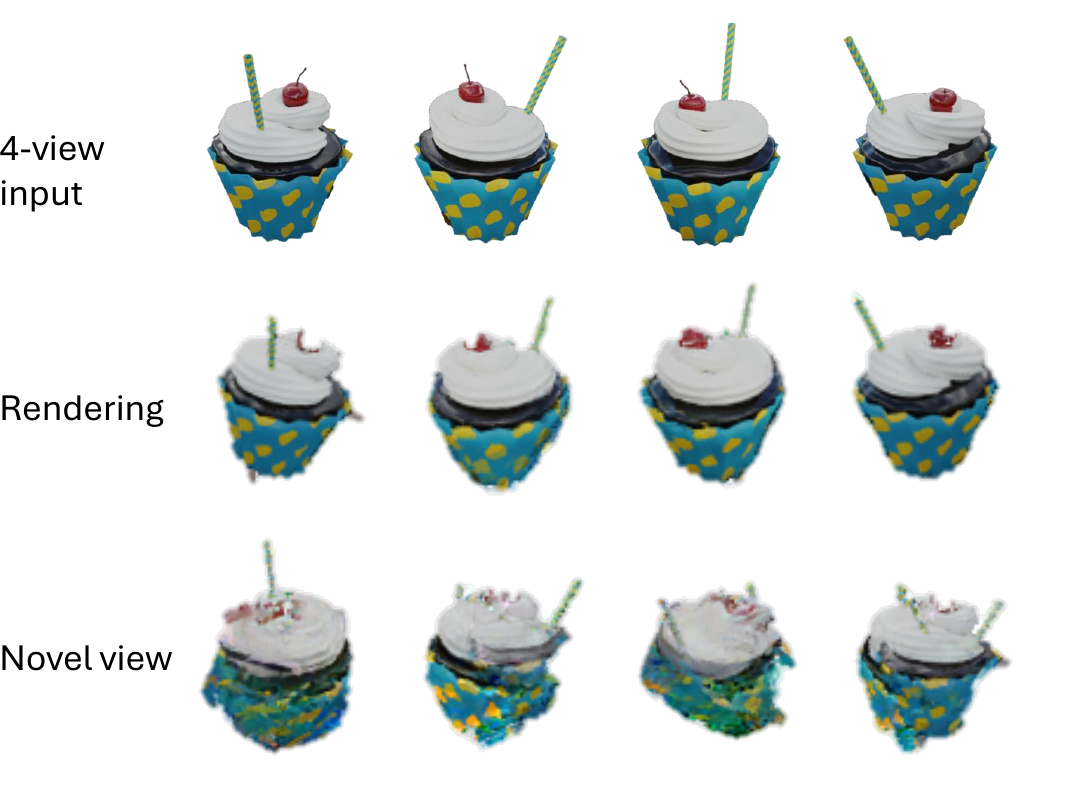}
    \caption{\textbf{GaussianObject results.} Due to the dependence on DuST3R, GaussianObject is still prone to errors. The rendering quality is worse than ours at novel views.}
    \label{fig:gsobject}
\end{figure}

\begin{table}[b]
\vspace{-0.1in}
\centering
\caption{\textbf{Reconstruction quality.} Compared with Nope-NeRF, our UVRM achieves higher reconstruction quality and is less fallible for 360-degree video input.}
\label{tab:recon_nope_nerf}
\vspace{-0.1in}
\resizebox{\linewidth}{!}{%
\begin{tabular}{cccc}
\toprule
    Model & PSNR ($\uparrow$)&SSIM ($\uparrow$)& Success rate\\\midrule
    Nope-NeRF~\cite{bian2023nope} &12.33&0.71&22\%\\
    UVRM &22.43&0.84&100\%\\\bottomrule
\end{tabular}
}
\vspace{-0.15in}

\end{table}
\section{Additional Results}
\subsection{Quantitative Comparison} 
The quantitative comparison results with Nope-NeRF are presented in \cref{tab:recon_nope_nerf}. Through experimentation, we observed that Nope-NeRF exhibits fragility to certain inputs, occasionally failing to reconstruct any reasonable shape. Consequently, we also include the success rate for Nope-NeRF. In comparison, our method demonstrates greater robustness and achieves superior reconstruction quality.
\subsection{Expanding Dataset Scale}
\Cref{fig:uvrm_128} demonstrates an additional experiment for our UVRM on 128 object collections. The training process takes around 3 days with 4 A100 GPUs using the same hyperparameters as in the main paper. As shown, UVRM, accompany with the proposed training method, is capable of reconstructing diverse objects concurrently. 
\subsection{Time Cost and Scalability}
Our proposed method yields increasing time efficiency as the number of objects grows. \Cref{fig:time_cost} illustrates this intuitively: as the size of object collections increases, the average reconstruction time per object decreases significantly. The per-object training time is lower than existing pose-free NeRF approaches (e.g., Nope-NeRF) when training with more than a collection of 20 objects concurrently. 
Besides, we also show that UVRM achieves better quantitative performance when scaling up.
All of these demonstrate the strong scalability potential of our approach.
\begin{figure}
    \centering
\includegraphics[width=.9\linewidth]{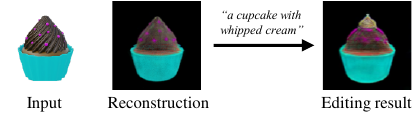}
    \caption{\textbf{Fast 3D editing.} Our training method achieves fast 3D editing due to the use of single reference image.}
    \label{fig:edit}
    \vspace{-0.1in}
\end{figure}
\subsection{Potential Applications.}
\noindent{\bf 3D editing.} 
Our pose-free training strategy can also be applied to edit a 3D object by propagating edits from a 2D reference view. As illustrated in \Cref{fig:edit}, given a reconstructed 3D object represented by a tri-plane, we can initially edit one reference frame using readily available text-to-image tools based on the user's prompt. Subsequently, we refine the initial 3D representation by employing our pose-free training strategy, specifically, the training objective detailed in Equation (12) of the main paper.

\subsection{Additional Ablation Experiments}
\begin{figure}
    \centering
    \includegraphics[width=.9\linewidth]{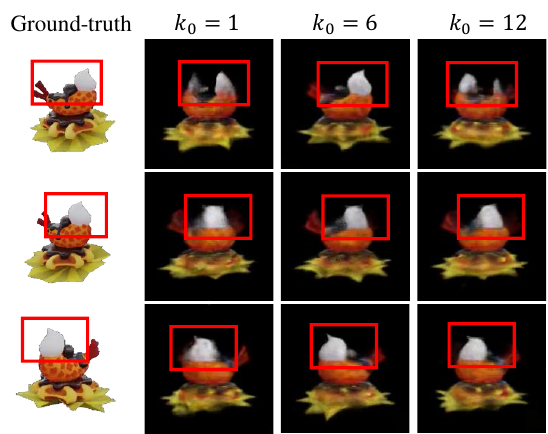}
    \caption{\textbf{Ablation study.} Qualitative comparison with various initial $k$ values.}
    \label{fig:ablation_k}
    \vspace{-0.1in}
\end{figure}
\noindent{\bf Hyper-parameters for Iterative Augmentation.} During pose-free training, we first synthesize $k$ pseudo-images and iteratively augment the pseudo-dataset. The initial value of $k$, denoted as $k_0$ here, is a crucial hyper-parameter, where higher $k_0$ results in view inconsistency due to the nature of Zero123-XL~\cite{liu2023zero1to3} (also shown in Fig. 7), and lower $k_0$ leads to weaker supervision. As an ablation study, we compare the results using different $k_0$ in Fig.~\ref{fig:ablation_k}. Based on our empirical experience, we recommend a $k_0$ value between 3 and 8.


\end{document}